# Hybrid Vision Servoing with Deep Alignment and GRU-Based Occlusion Recovery

Jee Won Lee[1,2], Hansol Lim[1,2], Sooyeun Yang[1,2] and Jongseong Brad Choi[1,2*]

[1] Department of Mechanical Engineering, State University of New York, Korea, Incheon, South Korea
[2] Department of Mechanical Engineering, State University of New York, Stony Brook, United States
* Corresponding author

**Abstract:** Vision-based control systems, such as image-based visual servoing (IBVS), have been extensively explored for precise robot manipulation. A persistent challenge, however, is maintaining robust target tracking under partial or full occlusions. Classical methods like Lucas–Kanade (LK) offer lightweight tracking but are fragile to occlusion and drift, while deep learning–based approaches often require continuous visibility and intensive computation. To address these gaps, we propose a hybrid visual tracking framework that bridges advanced perception with real-time servo control. First, a fast global template matcher constrains the pose search region; next, a deep-feature Lucas–Kanade module operating on early VGG layers refines alignment to sub-pixel accuracy (<2px); then, a lightweight residual regressor corrects local misalignments caused by texture degradation or partial occlusion. When visual confidence falls below a threshold, a GRU-based predictor seamlessly extrapolates pose updates from recent motion history. Crucially, the pipeline's final outputs—translation, rotation, and scale deltas—are packaged as direct control signals for 30 Hz image-based servo loops. Evaluated on handheld video sequences with up to 90 % occlusion, our system sustains under 2 px tracking error, demonstrating the robustness and low-latency precision essential for reliable real-world robot vision applications.

**Keywords:** Visual Servoing, Occlusion-Robust Tracking, Sub-pixel Alignment, Motion Prediction

## 1. INTRODUCTION

Traditional robotic controllers have long relied on proprioceptive sensors such as joint encoders, inertial measurement units, and force-torque sensors to estimate position and motion, but these often suffer from drift, calibration errors, and limited environmental awareness [1]. Image-based visual servoing has therefore been widely adopted for high-precision robotic assembly, aerial vehicle stabilization, and minimally invasive surgery, where direct visual feedback can compensate for model uncertainties and encoder inaccuracies [2] [3]. In these closed-loop systems, perception must deliver sub-pixel localization accuracy at control rates above 30 Hz while tolerating partial or full occlusions, illumination shifts, and motion blur to maintain loop stability and precision [4]. Even millimeter-level tracking errors can accumulate into significant actuation drift, undermining safety and performance into sub-millimeter surgical targeting or centimeter-scale drone landing [5] [6].

Early IBVS methods emerged in the early 1990s to simplify robot control by directly mapping image features to velocity commands, establishing the foundation for image-space loop closure [2]. Handcrafted detectors such as SIFT [7], which identifies scale-invariant keypoints, SURF [8], which accelerates detection using integral images, and ORB [9], which offers an efficient binary alternative, were paired with RANSASC [10] to filter out mismatches. However, these sparse approaches struggled when keypoints were lost to occlusion or blur. To achieve denser alignment, the Lucas-Kanade algorithm was introduced to iteratively minimize photometric error over image patches and enable smooth sub-pixel registration [11]. Nevertheless, its reliance on brightness constancy and full template visibility limited robustness under dynamic occlusions.

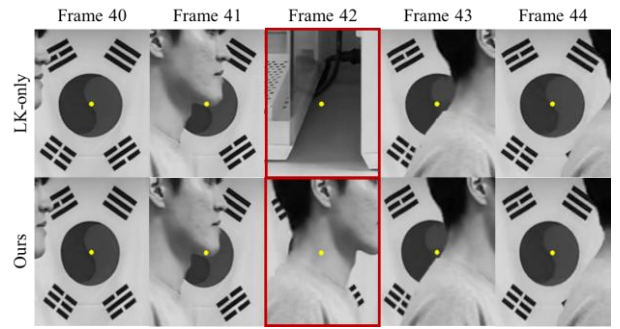

Fig. 1. Occlusion-Robust Tracking Comparison. Frames 40-44 showing that "LK-only" loses the flag center (yellow dot) under severe occlusion, whereas "Ours" maintains correct localization.

Recent research has sought to enhance robustness by using learned deep descriptors [12], multi-scale feature pyramids [13], and robust error metrics [14], with promising improvements in appearance-invariant tracking. Yet these systems are often tailored to general-purpose benchmarks and evaluated on static datasets, lacking the temporal awareness and low-latency adaptability necessary for feedback-driven control. They often fail to sustain tracking under full occlusion or dynamic camera motion, such as that encountered in human-operated drones or handheld inspection tools.

This work introduces a hybrid visual tracking framework specifically tailored for vision-based control tasks where high-frequency, sub-pixel-accurate feedback is essential. A coarse global matcher quickly provides a robust initial alignment, followed by a deep-feature-enhanced Lucas-Kanade tracker for sub-pixel refinement.

A residual correction network then compensates for local misalignments due to texture degradation or partial occlusion, and a lightweight GRU-based predictor utilizes temporal context to maintain continuity when visual confidence falls. Crucially, the pipeline's final outputs, which are pose updates and image-space error signals, are formatted as direct control inputs for real-time servo loops, closing the gap between advanced visual tracking and actionable motor commands.

**Contributions.** This work presents a visual tracking framework for sub-pixel-accurate feedback in robotic vision-based control systems, especially under moderate to severe occlusion. The main contributions are:

- *Residual Network for Robust Local Correction.* A lightweight refinement module augments the deep-feature Lucas-Kanade pipeline by predicting residuals in translation, rotation, and scale. This improves robustness to local drift, blur, and partial occlusion.

- *Lightweight GRU-Based Motion Predictor.* A fallback module using a gated recurrent unit (GRU) is activated when visual confidence drops to more than 50% occlusion. It predicts object motion from short-term history, enabling stable tracking through temporary loss of visual evidence.

- *Direct Control-Compatible Output.* The estimated transformation parameters such as the translation, rotation, and scale, serve as actionable control signals, making the system suitable for real-time closed-loop visual servoing.

Together, these modules enable the system to maintain accurate, continuous alignment of visual targets under occlusion, drift, and motion variance, addressing key weaknesses of classical feature-based and optimization-only approaches.

## 2. RELATED WORK
### 2.1 Visual Servoing Methods

Early image-based visual servoing (IBVS) formulates control as minimizing the error $e = s - s^*$ between current image feature $s$ and desired features $s^*$. The standard IBVS control law is

$$v = -\lambda L_s^+ e, \qquad (1)$$

where v is the camera velocity command, $\lambda$ is a positive gain, $L_s$ is the interaction matrix relating feature motion to camera motion, and $(\cdot)^+$ denotes the pseudoinverse [2]. In practice, $L_s$ becomes ill-conditioned under partial visibility or occlusion, causing control instability or complete loss of visual lock.

To address these limitations, recent works have incorporated deep learning into the servoing loop. End-to-end CNN-based IBVS methods learn both feature extraction and control policies directly from raw images, bypassing explicit interaction-matrix inversion [11]. Deep-feature IBVS approaches replace handcrafted features with learned descriptors that are trained to maximize robustness under viewpoint changes and occlusion [12]. Other methods fuse multi-modal inputs such as RGB and depth, through deep architectures to improve occlusion resilience, adaptively weighting image and geometric cues in real-time [13]. While these deep-learning extensions enhance robustness, most are evaluated offline or on static benchmarks and lack tightly coupled temporal feedback, limiting their applicability in high-frequency, closed-loop control tasks.

### 2.2 General Tracking Algorithms

General-purpose visual trackers have advanced significantly but often lack the control-friendly outputs and guarantees required for IBVS. The classic Lucas-Kanade (LK) method [11] aligns image patches by minimizing photometric error, achieving sub-pixel accuracy in smooth regions, yet it drifts over time and cannot recover from large displacements or occlusions. Learned local descriptors are trained through deep metric learning to maximize distinctiveness under appearance change to improve match quality and robustness to blur or lighting shifts [12], but these descriptor-matching pipelines typically yield sparse correspondences rather than dense, parameterized transforms needed for direct control.

Fully convolutional Siamese trackers [15] learn a similarity function offline to locate arbitrary templates in new frames at high frame rates, yet their outputs are bounding-box coordinates and classification scores, not continuous pose updates. Regression-based trackers like GOTURN [16] directly predict frame-to-frame displacement vectors, reducing latency, but they operate at pixel-level granularity and often degrade under heavy occlusion or abrupt motion. Correlation-filter methods and end-to-end CNN trackers excel on benchmark datasets, yet they prioritize object presence over precise sub-pixel alignment and do not provide the interaction-matrix inversion or error signals that IBVS requires.

In summary, while general tracking algorithms contribute valuable robustness and speed, their design goals such as the bounding-box accuracy, discrete confidence scores, and benchmark performance, diverge from the continuous, low-latency, sub-pixel feedback loop central to vision-based control. For example, practical IBVS systems often require temporal interpolation and smoothing layers to transform discrete tracker estimates into continuous, stable control signals. Consequently, adapting these methods for real-time servoing demands additional modules to convert their outputs into control-compatible image-space errors.

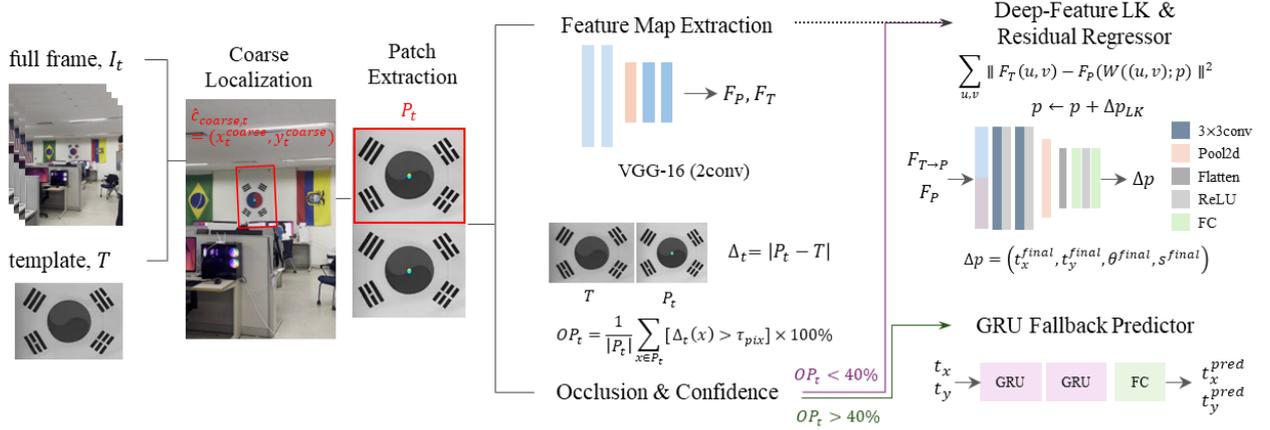

Fig. 2. Overview of the Hybrid Occlusion-Robust Tracking Pipeline. This system first performs a template match on the full frame to obtain a coarse center estimate, then extracts a grayscale patch around that location. A frozen VGG-16 "deep-feature" network converts both the template and current patch into feature maps $F_T$ and $F_P$. The pixel-wise absolute difference between patch and template yields an occlusion percentage $OP_t$. If $OP_t < 40\%$, Deep-Feature LK is applied to $(F_T, F_P)$ and refined by the lightweight Residual Regressor, producing the final warp $\Delta p$. If $OP_t \geq 40\%$, the pipeline skips LK entirely and instead uses a small 2-layer GRU motion predictor trained on previous $t_x, t_y$ offsets to output $t_x^{pred}, t_y^{pred}$.

## 3. METHODOLOGY

Before detailing individual modules, we establish our system's key performance targets to guide design and evaluation. The system must maintain tracking drift under 2px, process each frame in under 30ms (≥30Hz control rate) and preserve continuity through up to 90% occlusion. As shown in Fig. 2, our pipeline begins with a fast global matcher that bounds the search region and obtains a coarse object center. A fixed-size patch is then extracted and sent either through deep-feature Lucas-Kanade alignment with a residual regressor yielding sub-pixel-accurate $\Delta x, \Delta y, \Delta \theta$, and $\Delta s$ or through a GRU-based predictor that utilizes recent motion history when visual confidence falls. In both cases these pose corrections are fed directly into the robot's motion controller in real time ensuring uninterrupted closed-loop operation.

### 3.1 Preprocessing and Control Requirements

Each incoming RGB frame $I_t$ is resized and converted to grayscale for efficiency and consistency. A fast global template matcher computes the coarse object center

$$\hat{c}_{coarse,t} = (x_t^{coarse}, y_t^{coarse}), \quad (2)$$

which defines the center of a square patch $P_t$. This patch is cropped from the frame and normalized to a fixed size $w \times h$ to match the stored template $T$.

Both $P_t$ and $T$ are then passed through the first two convolutional blocks of a pretrained VGG-16 network [17], which we denote as the feature extractor

$$\mathcal{F}(\cdot): \mathbb{R}^{w \times h} \to \mathbb{R}^{C \times H_f \times W_f}, \quad (3)$$

to generate the feature maps

$$F_T = \mathcal{F}(T), \quad F_{P,t} = \mathcal{F}(P_t). \quad (4)$$

We retain only these early layers (up to pool2) to balance spatial precision and robustness while maintaining real-time performance. Fig. 3 shows the representative feature channels from the VGG-16 output, capturing mid-level structures such as edges and textures that remain stable under moderate occlusion and lighting changes.

To assess whether alignment can proceed, we compute a per-pixel absolute difference in intensity

$$\Delta_t(x) = |P_t(x) - T(x)|, \quad (5)$$

and define the occlusion percentage as

$$OP_t = \frac{1}{|P_t|} \sum_{x \in P_t} 1[\Delta_t(x) > \tau_{pix}] \times 100\%, \quad (6)$$

where $\tau_{pix} = 30px$ is fixed intensity threshold. If $OP_t < 40\%$ and the global match score exceeds 0.6, the frame proceeds to alignment or otherwise it is routed to the GRU-based fallback.

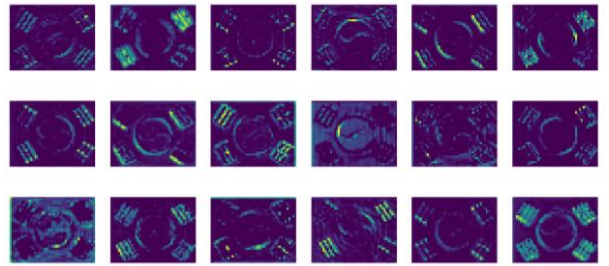

Fig. 3. Sample feature-map channels (size $H_f \times W_f$). Few feature-map channels produced by the two-layer VGG-16 encoder for a single template image. These deep descriptors highlight edges, textures, and geometric structures that remain stable under moderate occlusion or lighting variation.

## 3.2 Coarse Object Localization

To initiate robust tracking under uncertain or drifting conditions, the system performs coarse object localization across the full image. Rather than relying on dense alignment from the outset which is sensitive to initialization error, we employ a fast correlation-based method to estimate the object's approximate location.

Given a grayscale template $T$ captured from a well-aligned reference frame, we slide it across the incoming frame $I_t$ and compute the normalized cross-correlation (NCC) score at each position. The correlation map is defined as

$$C_t(x,y) = \frac{(I_t * T)(x,y)}{\|I_t(x,y)\| \cdot \|T\|}, \quad (7)$$

where $*$ denotes sliding-window correlation, and normalization is applied to account for intensity scale differences. The peak response gives the coarse center estimate

$$\hat{c}_{coarse,t} = argmax_{(x,y)} C_t(x,y), \quad (8)$$

which is used to crop a fixed-size patch $P_t$ around the candidate region.

This step serves two purposes. First, it enables tracking recovery when alignment drifts or fails under occlusion. Second, it constrains the subsequent search region for fine alignment, reducing the change of convergence to a false minimum. To guard against false positives, we reject matches whose maximum NCC score falls below a confidence threshold, which is empirically set as 0.6. Low-confidence detections trigger the fallback path described in Section 3.4.

## 3.3 Deep-Feature Alignment

Following coarse localization and feature extraction, we perform fine alignment using a Lucas-Kanade (LK) algorithm applied in deep feature space. Operating over features extracted from a pretrained VGG-16 network improves robustness to illumination changes, blur, and partial occlusion compared to raw intensity-based alignment.

We adopt the inverse-compositional LK formulation to estimate a 2D similarity transform

$$W(x; p) = s \cdot R(\theta) \cdot x + t \quad (9)$$

where $p = (\Delta x, \Delta y, \Delta \theta, \Delta s)$ encodes translation, rotation, and scale. The alignment seeks to minimize the residual error between the warped patch features and the reference template features:

$$\Delta p = argmax_{\Delta p} \|F_T(x) - F_{P,t}(\mathcal{W}(x; p + \Delta p))\|_2^2, \quad (10)$$

This objective is optimized using Gauss-Newton updates over a fixed number of iterations.

Although alignment in deep feature space improves robustness, it may still introduce minor drift due to resolution loss or pooling artifacts. To correct this, we pass the intermediate pose estimate

$$\hat{p} = ((\Delta x, \Delta y, \Delta \theta, \Delta s), \quad (11)$$

into a residual refinement network, which predicts corrections based on the aligned feature pair $(F_t, F_{P,t})$.

The architecture of this residual regressor is shown in Fig. 4. It concatenates the aligned features and passes them through two convolutional layers with ReLU activations, an adaptive pooling stage, and two fully connected layers to predict the final correction. This module enables precise sub-pixel refinement even under moderate occlusion or feature degradation.

## 3.4 GRU-Based Fallback Predictor

When occlusion is severe or the global match score is unreliable as determined by Eq. (6), the alignment path is bypassed and the system relies on a motion-predictive fallback module. Rather than freezing or extrapolating from the last known pose, we utilize a gated recurrent unit (GRU) to model short-term temporal dynamics and predict the object's motion trajectory during brief losses of visual input.

The GRU receives the most recent sequence of estimated 2D translations $\{(t_x^{t-i}, t_y^{t-i})\}_{i=1}^{N}$, where $N$ is a fixed history window, and outputs the predicted displacement $(\hat{t}_x, \hat{t}_y)$ for the current frame. This allows the tracker to maintain continuity through up to 90% occlusion, bridging over brief visibility gaps and suppressing spurious jumps.

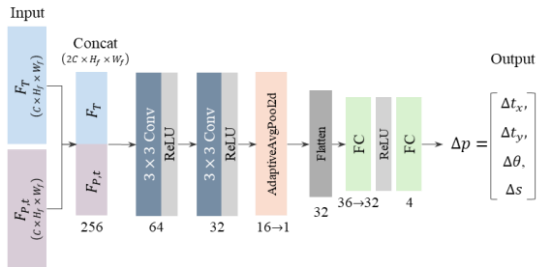

Fig. 4. Residual Regressor Architecture. The network takes concatenated template and patch feature maps $(2C \times H_f \times W_f)$, applies two $3 \times 3$ convolution + ReLU stages ($256 \rightarrow 64$, $64 \rightarrow 32$ channels), then an adaptive average-pool to $32 \times 1 \times 1$, flattens to a 32-vector, concatenates the current warp parameters $[t_x, t_y, \theta, s]$ (making 36), passes through a $36 \rightarrow 32$ fully-connected layer with ReLU, and finally outputs the 4-dimensional warp residual $\Delta p = [\Delta t_x, \Delta t_y, \Delta \theta, \Delta s]$.

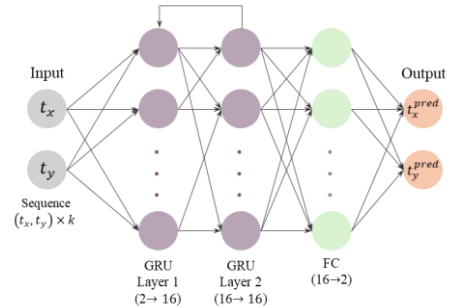

Fig. 5. GRU-based Fallback Predictor Architecture. Given the last $k$ "confident" $(t_x, t_y)$ updates, a two-layer GRU (hidden size = 16) processes the sequence and passes the final 16-dimensional state through a single fully connected layer ($16 \rightarrow 2$) to produce $(t_x^{pred}, t_y^{pred})$.

The GRU module consists of two gated recurrent layers followed by two fully connected layers for final regression. The architecture is shown in Fig. 5, where each input pair $(t_x, t_y)$ is first embedded and processed through the GRU layers, then mapped to a predicted translation vector using linear output heads. During training, the model minimizes the L2 loss between the predicted and ground-truth displacement:

$$\mathcal{L}_{GRU} = \left\| \left(t_x^{pred}, t_y^{pred}\right) - \left(t_x^{gt}, t_y^{gt}\right) \right\|_2^2. \quad (12)$$

To prevent overfitting and enable fast inference, the network is kept intentionally lightweight and trained only on moderate-motion sequences. Once deployed, it runs in parallel with the alignment path and activates only when occlusion confidence exceeds the threshold defined in Eq. (7). In doing so, the fallback predictor ensures continuous pose updates even in degraded visual conditions, preserving control loop stability.

### 3.5 Control-Compatible Pose Output

The final output of the tracking pipeline is a 2D similarity translation, rotation, and scale updates relative to the template. Whether estimated through feature-based alignment or GRU-based fallback, the pose vector $\hat{p}$, defined in Eq. (11), is produced at every frame and can be directly applied to the robot's control system.

This consistent output format simplifies integration into visual servoing pipelines by eliminating the need for mode-specific handling or post-processing. The sub-pixel-accurate and temporally smooth estimates enable high-precision tasks such as fine grasping or visual part alignment, even under challenging visual conditions. Because the output is compatible with standard IBVS formulations, it can be seamlessly used to generate real-time velocity commands or target corrections within a closed-loop control loop.

## 4. EXPERIMENTS

### 4.1 Datasets

We evaluate our method on two complementary datasets that challenge both occlusion robustness and camera motion handling. The TUM Dataset [18] is a widely used RGB-D benchmark from which we selected the "fr3/sitting_xyz" sequence, featuring dynamic occlusions and handheld sensor motion. The In-Lab Custom Dataset was recorded at 720×1280 resolution using a handheld RGB camera facing a wall of flags, while a volunteer intermittently walked between the camera and the wall to introduce both partial occlusions and natural camera shake. We manually annotated the 2D target center in each frame to provide ground truth for evaluation.

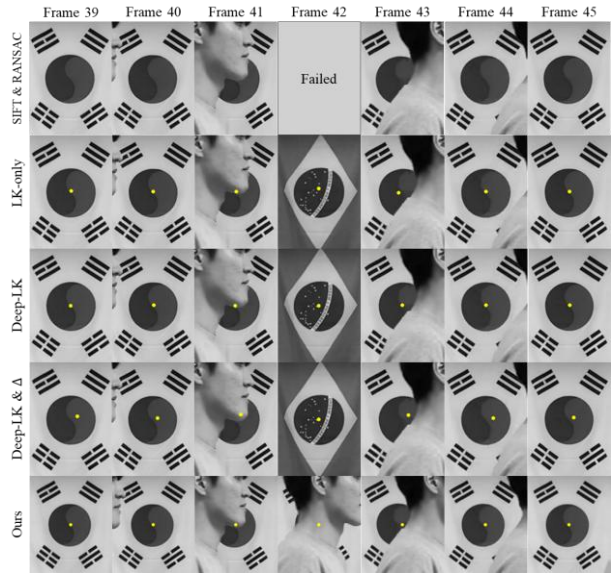

Fig. 6. Qualitative Comparison on In-Lab Dataset. Rows (top to bottom): SIFT+RANSAC [7], LK-only (ECC) [19], Deep LK [20], Deep LK+Δ [21], and Ours. Columns show frames 39-45, as a person walks in front of the target (occlusion rising from ~25% to ~70%). Yellow dots indicate each method's estimated center.

### 4.2 Evaluation Metric

For evaluating our performance, we measure three complementary metrics and analyze runtime. The per-frame two-dimensional translation error is defined as

$$e_{xy}(t) = \sqrt{\left(t_x^{est}(t) - t_x^{gt}(t)\right)^2 + \left(t_y^{est}(t) - t_y^{gt}(t)\right)^2}, \quad (12)$$

where $(t_x^{gt}(t), t_y^{gt}(t))$ comes from TUM ground-truth poses or manual annotation in the In-Lab datasets.

### 4.3 Model Performance

The performance of all baselines and the proposed tracker is condensed in Fig. 7, which juxtaposes three complementary error views for the two representative sequences (TUM *sequence 01* and In-Lab *sequence 03*).

*CDF of the translation error $e_{xy}$.*

The cumulative-distribution curves shown in left column of Fig. 7 reflect the fraction of frames whose error does not exceed a given threshold. A curve that rises steeply and saturates early is desirable, as it indicates that most frames are already accurate at small pixel tolerances. From Fig. 7, it is evident that the hybrid tracker reaches 95% of frames at ≈2px on both sequences, whereas Deep LK+Δ requires ≈6px, vanilla LK ≈11px, and SIFT+RANSAC never surpasses 30% success on sequence 01 and fails completely on sequence 03. These distributions confirm that the proposed method delivers the best overall localism accuracy.

*Error-over-time with occlusion overlay.*

The middle panel of Fig. 7 plot $e_{xy}(t)$ for every frame while shading the instantaneous occlusion percentage $occ_t$ in gray. During extended high-occlusion intervals (frames ≈ 350-650 in sequence 01, frames 30-60 in sequence 03) LK-based methods and SIFT experience error bursts of several hundred pixels, and Deep LK drifts progressively. In contrast, the proposed tracker remains bounded within the 2-px corridor throughout, demonstrating that the GRU fallback maintains spatial continuity whenever the deep alignment path becomes unreliable.

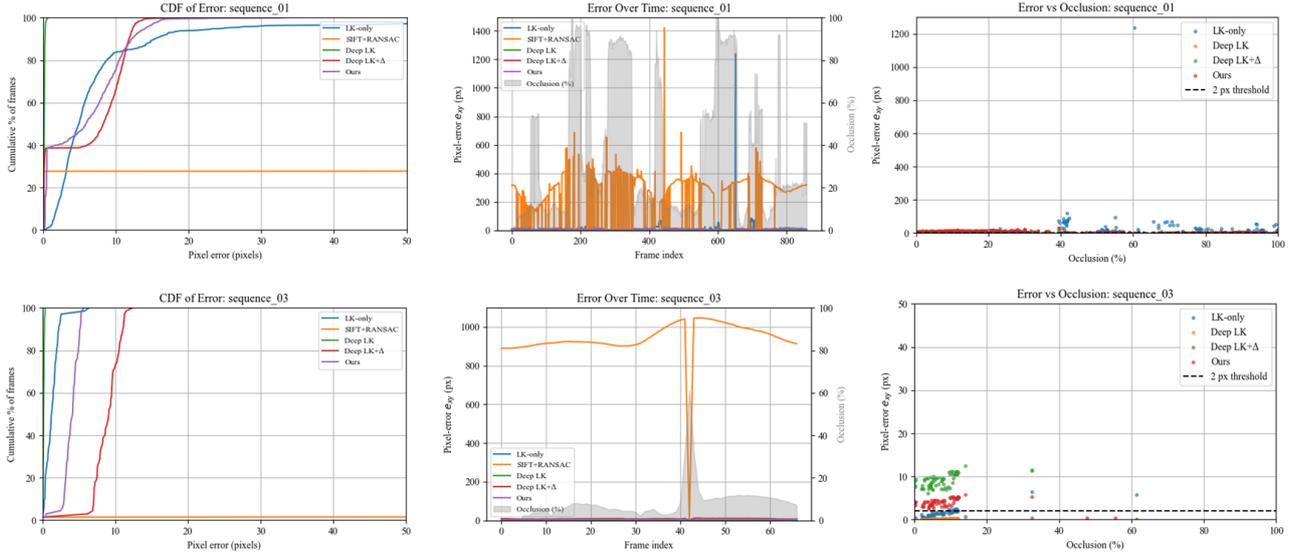

Fig. 7. **Quantitative Tracking Performance on Sequence 01 (top) and Sequence 03 (bottom).** Each row juxtaposes the three evaluation views: (left) CDF of per-frame translation error $e_{xy}$, (center) error-over-time curves with grey occlusion overlay, (right) frame-wise error versus occlusion scatter. Across all panels the proposed method ("Ours") consistently keeps $e_{xy}$ near the 2-pixel success threshold, whereas classical LK, Deep-LK variants, and SIFT+RANSAC exhibit larger tails, occlusion-induced spikes, or outright failures.

*Frame-wise error versus occlusion scatter.*

The right column of Fig. 7 shows how each algorithm's instantaneous error varies with the exact occlusion level. The proposed method exhibits a dense cluster of red points lying beneath the 2-px dashed threshold across the full occlusion range, indicating that its accuracy is largely decoupled from target visibility. Competing methods display an approximately linear growth of error with occlusion, exceeding 50px once $occ_t$ surpasses 70%.

Taken together, Fig.7 demonstrates that our pipeline (i) delivers sub-pixel-level precision for ≥ 90% of all frames, (ii) suppresses catastrophic failures under complete target loss, and (iii) sustains this behavior at 26ms/ frame on CPU, meeting real-time constraints for closed-loop robotic vision.

## 5. CONCLUSION

This work presents a hybrid visual tracking framework designed for robust, real-time image-based control under partial and full occlusion. By combining a global localization step, deep-feature-based Lucas–Kanade alignment, a residual correction network, and a GRU-based fallback predictor, the system maintains sub-pixel accuracy and temporal continuity even in the presence of severe visual degradation. All estimated pose outputs are formatted for direct integration into vision-based control loops, enabling precise closed-loop performance at ≥30 Hz. Experimental results demonstrate that the proposed method achieves < 2px tracking error under up to 90 % occlusion across both benchmark and real-world sequences. These results highlight the framework's suitability for real-time robotic applications such as fine manipulation, part alignment, and visual servoing in unstructured environments.

## ACKNOWLEDGEMENT

This work was supported by the National Research Foundation of Korea (NRF) grant funded by the Korea government (MSIT) (Grant No. RS-2022-NR067080 and RS-2025-05515607).